\documentclass[11pt]{article}

\usepackage[preprint]{acl}

\usepackage{times}
\usepackage{latexsym}

\usepackage[T1]{fontenc}

\usepackage[utf8]{inputenc}

\usepackage{microtype}

\usepackage{inconsolata}

\usepackage{graphicx}
\usepackage{booktabs}
\usepackage{amssymb}
\usepackage{algorithm}
\usepackage{algpseudocode}
\usepackage{multirow}
\usepackage{xcolor}
\usepackage{tcolorbox}
\definecolor{gainup}{RGB}{0,140,60}
\definecolor{gaindown}{RGB}{200,30,30}
\newcommand{\up}[1]{\textsubscript{\textcolor{gainup}{$\uparrow$#1}}}
\newcommand{\dn}[1]{\textsubscript{\textcolor{gaindown}{$\downarrow$#1}}}
\newcommand{\nc}[1]{\textsubscript{\textcolor{gray}{#1}}}

\tcbuselibrary{most}
\newtcolorbox{skillbox}[2][]{breakable, enhanced, colback=gray!6,
  colframe=black!75, coltitle=white, fonttitle=\bfseries\small, title={#2},
  boxrule=0.6pt, arc=1.5mm, left=2mm, right=2mm, top=1.5mm, bottom=1.5mm,
  fontupper=\small, before skip=8pt, after skip=8pt,
  segmentation style={solid, black!45}, #1}
\makeatletter
\newcommand{\compactlists}{%
  \setlength{\leftmargini}{1.5em}\setlength{\leftmarginii}{1.2em}%
  \def\@listi{\leftmargin\leftmargini \topsep 1pt \parsep 0pt \itemsep 0pt}%
  \let\@listI\@listi
  \def\@listii{\leftmargin\leftmarginii \topsep 0pt \parsep 0pt \itemsep 0pt}}
\makeatother
\tcbset{skillfloat/.style={unbreakable, fontupper=\footnotesize,
  before skip=0pt, after skip=0pt,
  before upper={\compactlists\setlength{\parskip}{2pt}}}}

\title{SkillACE: Fine-Grained Agent Skill Evolution \\ via an Adaptive Error Taxonomy}

\author{
  {\bfseries Kangcheng Luo\textsuperscript{\normalfont 1}\thanks{Equal contribution.}, Weijia Zhang\textsuperscript{\normalfont 2}\footnotemark[1], Jue Zhang\textsuperscript{\normalfont 1}, Yiming Jia, Zhe Li\textsuperscript{\normalfont 1},} \\
  {\bfseries Xuzheng Chen\textsuperscript{\normalfont 1}, Chenglei Shen, Shuheng Cao, Hongyi Du,} \\
  {\bfseries Yijie Lu\textsuperscript{\normalfont 1}, Xingyuan Lu, Zhenwei An, Kun Xu, Yansong Feng\textsuperscript{\normalfont 1}\thanks{Corresponding author.}} \\
  \textsuperscript{1}Peking University \quad \textsuperscript{2}Yale University \\
  \texttt{luokangcheng@stu.pku.edu.cn} \quad
  \texttt{fengyansong@pku.edu.cn}}

\begin{document}
\maketitle
\begin{abstract}
Skill self-evolution enables large language model (LLM) agents to accumulate
reusable procedural knowledge from execution experience without updating
parameters. Existing methods either consolidate this experience into a single
shared Skill or maintain a library of Skills, each revised with feedback from
the tasks that invoke it. Tasks sharing a Skill, however, often fail for
different reasons, so each revision mixes unrelated errors and a rule that
fixes one error interferes with other tasks.
We argue that error categories are the natural unit for organizing and
revising Skills, because failures that share an error call for the same
correction. Since these errors vary across environments and only surface
during execution, we propose \textbf{SkillACE}, which induces an adaptive error
taxonomy from observed failures and pairs each category with one fine-grained
Skill revised only with that category's failures. Because this pairing
persists throughout evolution, the correction for each error accumulates in its
own Skill. On four agentic benchmarks, SkillACE outperforms all compared baselines,
improving the average score by 11.5 points over using no Skill, and its Skills
transfer effectively to other LLMs. These results establish the adaptive error taxonomy
as an effective unit for accumulating agent experience.
\end{abstract}

\section{Introduction}
\label{sec:introduction}

Large language model (LLM) agents can complete complex, long-horizon tasks in
interactive environments \citep{dong2026longhorizon}, and their execution
experience can improve performance on subsequent unseen tasks. Skills store
this experience as reusable documents, allowing agents to
accumulate procedural knowledge without updating model parameters
\citep{ni2026trace2skill,yang2026skillopt,mi2026skillpro,qiu2026autorefine}.
As experience accumulates across tasks, a central design choice is the unit of
accumulation, which decides both how Skills are organized and which failures
revise each Skill.

Existing methods either consolidate the experience from all tasks into a
single shared Skill
\citep{ni2026trace2skill,yang2026skillopt,chen2026skillcat} or maintain a
library of Skills, each revised with feedback from the tasks that invoke it
\citep{mi2026skillpro,zhou2026mementoskills,ma2026skillclaw,lin2026museautoskill}.
However, tasks covered by the same Skill often fail for different reasons, so
each revision of that Skill is driven by a mixture of unrelated errors. This
leads to rule interference: a rule added to fix one error interferes with
other tasks covered by the same Skill, breaking tasks that the agent could
already solve (Appendix~\ref{sec:case-study}). A few methods categorize
failures, but their categories are predefined and coarse
\citep{liu2026skillforge} or are used only to synthesize new Skills, which
are never revised afterward \citep{zhang2026ratchet}.

We argue that error categories are the natural unit for both organizing and
revising Skills. Failures that share an error stem from the same deficiency in
the agent's actions or judgments, and therefore call for the same procedural
correction \citep{zhu2025agentdebug,cemri2026adamast}. Error categories are
finer than the set of tasks that invoke a Skill, so a correction for one error
does not interfere with tasks that fail for other reasons. They are also
coarser than individual trajectories, so each revision draws on failures that
recur across tasks and is less tied to the incidental details of a single
trajectory. The errors an agent makes, however, vary across environments and
only surface during execution, so the error taxonomy must be induced from
failures observed during evolution.

\begin{figure*}[t!]
    \centering
    \includegraphics[width=\textwidth]{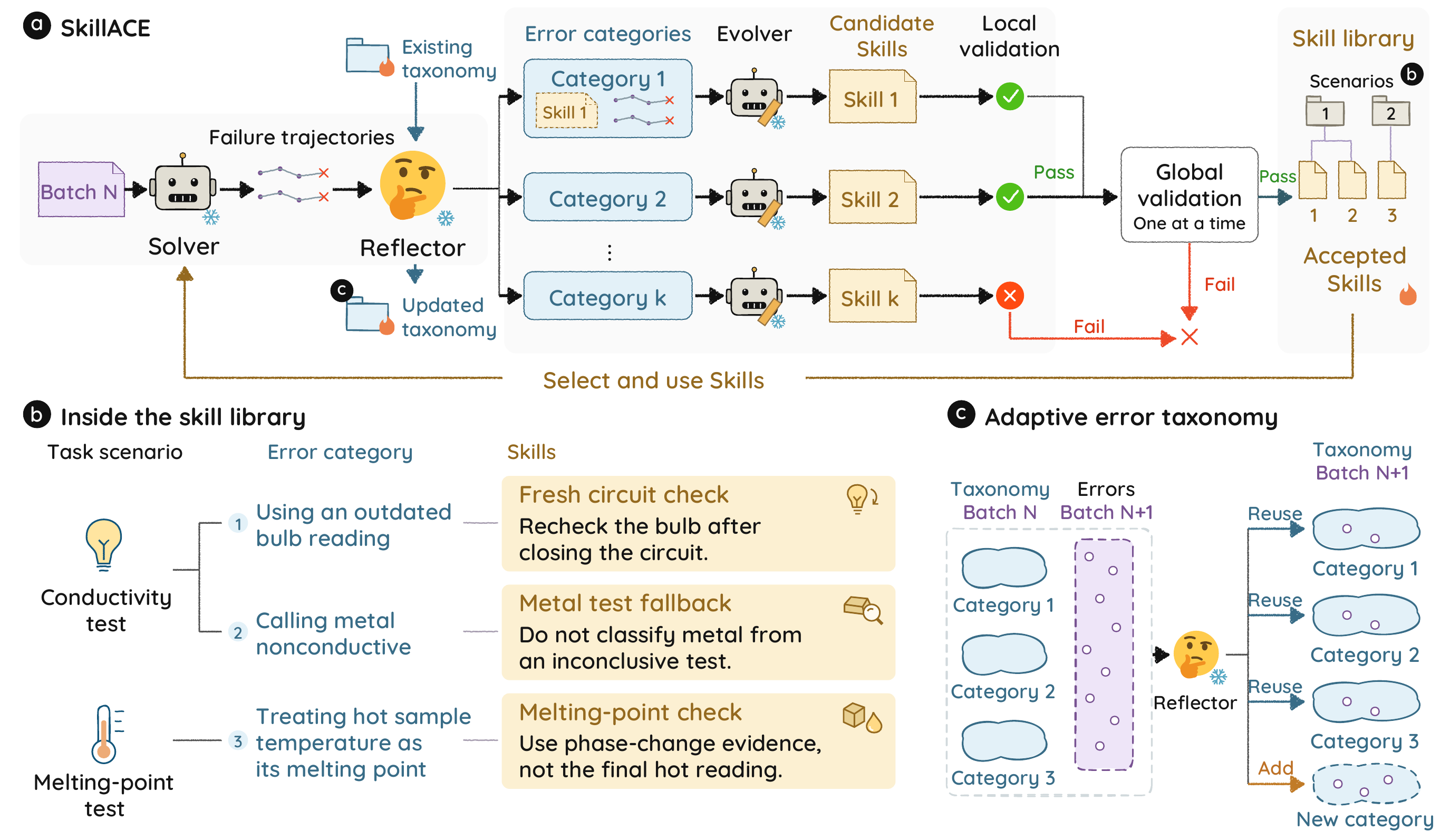}
    \caption{Overview of SkillACE. (a) In each batch, the Solver executes tasks, the Reflector categorizes failures, and the Evolver updates target Skills, which enter the library after local and global validation. (b) Inside the library, each error category is paired with one fine-grained Skill. (c) Throughout evolution, the taxonomy reuses existing categories and adds new ones, so recurring failures keep refining the same Skill.}
    \label{fig:method-overview}
\end{figure*}

We therefore propose \textbf{SkillACE} (Adaptive Category-guided Evolution),
which organizes and revises Skills through an evolving error taxonomy
(Figure~\ref{fig:method-overview}). In each batch, the Solver executes tasks,
the Reflector assigns each failure to an existing error category or creates a
new category for it, and the Evolver revises the Skill of each selected
category using only failures of that category. A revised Skill enters the
library only after passing local and global validation. Each error category is
paired with one fine-grained Skill, and this pairing is kept throughout
evolution. When an error recurs later, its failures are assigned to the
existing category and revise the same Skill, so the correction for each error
accumulates in a single Skill.

We evaluate SkillACE on four agentic benchmarks spanning embodied interaction
\citep{wang2022scienceworld,shridhar2021alfworld}, web shopping
\citep{yao2022webshop}, and spreadsheet manipulation
\citep{ma2024spreadsheetbench}. With Qwen3.6-35B-A3B \citep{qwen2026qwen36}
serving as the Solver, Reflector, and Evolver, SkillACE achieves the best
result on every benchmark, improving the average score by 11.5 points over
using no Skill. The learned Skills also transfer effectively to Qwen3.5-9B
\citep{qwen2026qwen35} and Gemma-4-12B-it \citep{gemmateam2026gemma4}
without re-evolution.

Our contributions are as follows:
(1) \textbf{An error-centric view of Skill evolution.} We shift Skill evolution
from task-centric to error-centric organization: experience is grouped by the
errors the agent makes, so that each Skill captures the correction for one
error and is shielded from revisions aimed at others.
(2) \textbf{Taxonomy-guided Skill evolution.} SkillACE grows an error taxonomy
from the agent's own failures and binds each category to one fine-grained
Skill throughout evolution, turning recurring failures into cumulative
refinement of the same Skill.
(3) \textbf{Strong and transferable gains.} Experiments on four agentic benchmarks
show that SkillACE achieves the best results and that the evolved Skills
transfer effectively to other LLMs without re-evolution.

\section{Related Work}
\label{sec:related-work}

\paragraph{Learning and Evolving Skills from Experience.}
Agents learn from execution feedback by retaining past experiences in memory
\citep{shinn2023reflexion,zhao2023expel,wang2023voyager,fang2025memp} or by
optimizing their prompts and other components
\citep{zhou2024agentsymboliclearning,agrawal2025gepa}. Skills are a form of such
memory that stores procedural knowledge as reusable instructions. They are
constructed from trajectories
\citep{ni2026trace2skill,wang2026skillx} or user interactions
\citep{yang2026autoskill,ma2026skillclaw} and refined with execution feedback,
validation, or training
\citep{zhang2026skillevolver,qiu2026autorefine,zhang2026coevoskills,yang2026skillopt,xia2026skillrl,ouyang2026skillos}.
These methods differ in the unit of accumulation. Some consolidate the
experience from all tasks into a single shared Skill
\citep{ni2026trace2skill,yang2026skillopt,chen2026skillcat}, while others
maintain a library of Skills and revise each with feedback from the tasks that
invoke it
\citep{mi2026skillpro,zhou2026mementoskills,ma2026skillclaw,lin2026museautoskill}.
In both designs the unit is defined by tasks, so each Skill is revised by
failures with different causes. SkillACE instead makes the error category the
unit of accumulation, so each Skill is revised only by failures of one
category.

\paragraph{Failure Diagnosis for Skill Revision.}
Failure diagnosis connects execution failures to their causes. AgentDebug
locates critical errors within a trajectory to guide re-execution
\citep{zhu2025agentdebug}, and AdaMAST induces error categories across runs as
feedback for improving agent systems \citep{cemri2026adamast}. Their diagnoses,
however, do not determine which knowledge a failure should revise. Some
methods group failures to guide revision, but in limited ways: SkillForge
relies on predefined dimensions \citep{liu2026skillforge}, Ratchet
\citep{zhang2026ratchet} uses failure clusters only to synthesize new Skills,
never revising existing ones, and MNL \citep{su2026mnl} clusters each batch by
the subject of the question instead of by error and merges the distilled
notes into memory entries matched by embedding similarity. In SkillACE,
categories are induced from the errors behind observed failures and expanded
as new errors emerge, and a recurring error is assigned to its existing
category, so its later failures keep revising the same Skill.

\section{Method}
\label{sec:method}

\subsection{Problem Setup}
\label{sec:problem-setup}

We consider an LLM agent with frozen parameters. Its behavior is improved only
through a library of Skill documents $\mathcal S=\{s_1,\dots,s_n\}$. Skills
are loaded through progressive disclosure. When executing a task $x$, the agent
sees only the summary and applicable scenario of each Skill, and loads a full
Skill into context only when relevant. Each execution produces a trajectory $\tau$
and a score $r$:
\begin{equation}
(\tau, r) = h(M, x, \mathcal S), \qquad r \in [0, 1],
\end{equation}
where $M$ is the frozen model and $h$ is the agent harness that runs $M$ with
$\mathcal S$ on task $x$ in the environment.

The tasks of each benchmark are split into three disjoint sets with distinct roles: the evolution
set $D_{\mathrm{evolve}}$ supplies experience, the validation set
$D_{\mathrm{val}}$ decides whether new or revised Skills are accepted, and the
test set $D_{\mathrm{test}}$ is used only for final reporting. Let
$r(x; \mathcal S)$ denote the score on task $x$ under library $\mathcal S$ and
$\mathcal D$ the task distribution of the benchmark. Using only
$D_{\mathrm{evolve}}$ and $D_{\mathrm{val}}$, the goal is to find
\begin{equation}
\mathcal S^\star = \arg\max_{\mathcal S} \;
\mathbb{E}_{x \sim \mathcal D}\big[r(x; \mathcal S)\big],
\end{equation}
whose performance is measured on the held-out $D_{\mathrm{test}}$.
Evolution processes $D_{\mathrm{evolve}}$ in successive batches. At batch $t$, the agent executes
tasks $B_t \subseteq D_{\mathrm{evolve}}$ with the current library
$\mathcal S_t$, and the resulting trajectories and feedback are used to update
the library to $\mathcal S_{t+1}$ for the next batch. Let $F_t$
denote the trajectories of the tasks that fail in batch $t$.

\subsection{Framework Overview}
\label{sec:framework-overview}

SkillACE uses the error category as the unit of experience
accumulation: each error category corresponds to one Skill, which is revised
only with failures of that category. As shown in
Figure~\ref{fig:method-overview}, each batch is processed by three components
in turn. The Solver executes $B_t$ with the current library $\mathcal S_t$. The
Reflector reads the failure trajectories $F_t$ and assigns each failure
to an existing category or creates a new category for a newly emerging error,
expanding the error taxonomy batch by batch
(Section~\ref{sec:taxonomy-construction}). The Evolver revises or creates the
Skills of the selected categories (Section~\ref{sec:skill-evolution}). A
revised Skill must pass local validation on tasks of its category and then
global validation on the validation set before entering
$\mathcal S_{t+1}$.

\subsection{Adaptive Error Taxonomy Construction}
\label{sec:taxonomy-construction}

\begin{algorithm}[!t]
\caption{SkillACE: Skill Evolution Guided by an Adaptive Error Taxonomy}
\label{alg:skillace}
\small
\algrenewcommand\algorithmicindent{0.9em}
\algrenewcommand\algorithmicrequire{\textbf{Input:}}
\algrenewcommand\algorithmicensure{\textbf{Output:}}
\begin{algorithmic}[1]
\Require evolution set $D_{\mathrm{evolve}}$, validation set $D_{\mathrm{val}}$,
number of categories $K$, cycle budget $R$, number of epochs $E$
\Ensure evolved Skill library
\State $\mathcal S_1 \gets \emptyset$; $\phi_1 \gets \emptyset$; $t \gets 1$
\State $\mathcal T_1 \gets$ taxonomy with no error categories
\For{$e = 1, \dots, E$}
\For{each batch $B_t$ of $D_{\mathrm{evolve}}$}
  \State $F_t \gets \textsc{Solver}(B_t, \mathcal S_t)$ \Comment{failure trajectories}
  \State $(\mathcal T_{t+1}, \{F_t^c\}, \phi_t') \gets \textsc{Reflector}(F_t, \mathcal T_t, \phi_t)$
  \State $\mathcal C_{\mathrm{sel}} \gets$ the $K$ categories with the largest $|F_t^c|$
  \State $P \gets \emptyset$ \Comment{locally validated candidates}
  \For{each $c \in \mathcal C_{\mathrm{sel}}$}
    \State $\mathit{feedback} \gets \varnothing$ \Comment{from local validation}
    \For{$r = 1, \dots, R$}
      \State $s_c' \gets \textsc{Evolver}(F_t^c, \phi_t'(c), \mathit{feedback})$
      \If{$\mathcal S_t \oplus s_c'$ passes local validation}
        \State $P \gets P \cup \{s_c'\}$; \textbf{break}
      \EndIf
      \State $\mathit{feedback} \gets$ unsolved and regressed tasks
    \EndFor
  \EndFor
  \State $\mathcal S \gets \mathcal S_t$; $\phi \gets \phi_t'$
  \For{each $s_c' \in P$ in order}
    \If{$\mathcal S \oplus s_c'$ passes global validation on $D_{\mathrm{val}}$}
      \State $\mathcal S \gets \mathcal S \oplus s_c'$; $\phi(c) \gets s_c'$
    \EndIf
  \EndFor
  \State $\mathcal S_{t+1} \gets \mathcal S$; $\phi_{t+1} \gets \phi$; $t \gets t + 1$
\EndFor
\EndFor
\State \Return $\mathcal S_t$
\end{algorithmic}
\end{algorithm}

SkillACE builds the error taxonomy from scratch, starting with no error
categories and inducing every category from failures observed during
evolution. The taxonomy has two levels (Figure~\ref{fig:method-overview}b).
The first level consists of task scenarios, such as the scientific
topics of Mendelian genetics and electrical circuits in ScienceWorld
\citep{wang2022scienceworld}. The second level consists of error
categories. Each category $c$ belongs to one scenario and carries a description
$d_c$ that characterizes one erroneous agent behavior in tasks of that
scenario, such as "deciding genetic dominance without sufficient inheritance
evidence under Mendelian genetics". We write $\mathcal T_t$ for the taxonomy at
the start of batch $t$ and $\mathcal C_t$ for its error categories.
Because the errors of an agent are hard to enumerate in
advance, $\mathcal C_1$ is empty. Initial scenarios can optionally be taken
from the task types provided by a benchmark, such as the 10 scientific topics
in ScienceWorld and cell-level and sheet-level manipulation in SpreadsheetBench
\citep{ma2024spreadsheetbench}. Otherwise, the Reflector creates scenarios as
failures arrive.

In batch $t$, the Reflector expands the taxonomy with the failures $F_t$. For
each failure, it reads the Solver's trajectory and writes a reflection on the
erroneous behavior that caused the failure. It then checks whether an existing
category already covers this error, based on the category descriptions and
the content of their Skill. A failure covered by an existing category is assigned to
it, and a newly emerging error forms a new category under its scenario. When
no existing scenario fits, the Reflector also adds a scenario, and it may
revise or merge existing categories and scenarios. Each
failure trajectory $\tau \in F_t$ is assigned to exactly one category
$c_t(\tau) \in \mathcal C_{t+1}$, corresponding to the primary error that directly
blocked task completion. Secondary errors in the same trajectory are not
discarded. Once an accepted Skill revision fixes the primary error, a task that
still fails in a later epoch exposes its next error as the primary one, which
is then categorized in turn. Each update thus targets a single error. We write
$F_t^c = \{ \tau \in F_t : c_t(\tau) = c \}$ for the failures of category $c$. The
updated taxonomy $\mathcal T_{t+1}$ is used for categorization in subsequent
batches, regardless of whether the Skill revisions of the current batch pass
validation.

\begin{table*}[t]
    \centering
    \setlength{\tabcolsep}{3pt}
    \renewcommand{\arraystretch}{1.08}
    \resizebox{\textwidth}{!}{%
    \begin{tabular}{@{}*{12}{c}@{}}
        \toprule
        Solver & Method
        & \multicolumn{5}{c}{Reflector \& Evolver: GPT-5.5}
        & \multicolumn{5}{c}{Reflector \& Evolver: Qwen3.6-35B-A3B} \\
        \cmidrule(lr){3-7} \cmidrule(l){8-12}
        & & ScienceWorld & ALFWorld & Spreadsheet & WebShop & Avg
        & ScienceWorld & ALFWorld & Spreadsheet & WebShop & Avg \\
        \midrule
        \multirow{7}{*}{\shortstack[c]{Qwen3.6-\\35B-A3B}} & No Skill & 59.3 & 93.8 & 61.5 & 32.6 & 61.8 & 59.3 & 93.8 & 61.5 & 32.6 & 61.8 \\
         & LLMGen & 64.6\up{5.3} & 95.8\up{2.0} & 62.7\up{1.2} & \underline{38.1}\up{5.5} & 65.3\up{3.5} & 65.8\up{6.5} & 93.0\dn{0.8} & 61.0\dn{0.5} & 34.5\up{1.9} & 63.6\up{1.8} \\
        \cmidrule(l){2-12}
         & SkillOpt & \underline{82.3}\up{23.0} & 95.3\up{1.5} & 63.9\up{2.4} & 37.9\up{5.3} & \underline{69.9}\up{8.1} & \underline{66.3}\up{7.0} & \underline{97.0}\up{3.2} & \underline{68.3}\up{6.8} & \underline{38.7}\up{6.1} & \underline{67.6}\up{5.8} \\
         & AutoRefine & 51.8\dn{7.5} & 94.3\up{0.5} & 58.7\dn{2.8} & 37.3\up{4.7} & 60.5\dn{1.3} & 54.5\dn{4.8} & 89.1\dn{4.7} & 56.7\dn{4.8} & 34.3\up{1.7} & 58.7\dn{3.2} \\
         & Trace2Skill & 71.3\up{12.0} & \underline{97.8}\up{4.0} & \underline{68.7}\up{7.2} & 33.6\up{1.0} & 67.9\up{6.1} & 60.0\up{0.7} & 93.8\nc{$\pm$0.0} & 59.9\dn{1.6} & 33.2\up{0.6} & 61.7\dn{0.1} \\
         & Skill-Pro & 51.1\dn{8.2} & 89.8\dn{4.0} & 54.6\dn{6.9} & 37.1\up{4.5} & 58.2\dn{3.6} & 53.9\dn{5.4} & 88.6\dn{5.2} & 55.1\dn{6.4} & 36.1\up{3.5} & 58.4\dn{3.4} \\
        & \textbf{SkillACE} & \textbf{84.9}\up{25.6} & \textbf{99.5}\up{5.7} & \textbf{82.3}\up{20.8} & \textbf{38.7}\up{6.1} & \textbf{76.4}\up{14.6} & \textbf{78.9}\up{19.6} & \textbf{99.5}\up{5.7} & \textbf{74.5}\up{13.0} & \textbf{40.1}\up{7.5} & \textbf{73.3}\up{11.5} \\
        \midrule
        \multicolumn{12}{c}{\textit{Cross-Model Transfer}} \\
        \midrule
        \multirow{7}{*}{\shortstack[c]{Qwen3.5-\\9B}} & No Skill & 47.3 & 50.5 & 42.1 & 21.9 & 40.5 & 47.3 & 50.5 & 42.1 & 21.9 & 40.5 \\
         & LLMGen & 52.1\up{4.8} & \underline{84.1}\up{33.6} & 41.4\dn{0.7} & 28.3\up{6.4} & 51.5\up{11.0} & 47.0\dn{0.3} & 66.4\up{15.9} & 41.5\dn{0.6} & 26.5\up{4.6} & 45.4\up{4.9} \\
        \cmidrule(l){2-12}
         & SkillOpt & \underline{65.4}\up{18.1} & 76.6\up{26.1} & \underline{45.1}\up{3.0} & 28.9\up{7.0} & \underline{54.0}\up{13.6} & \underline{53.4}\up{6.1} & 70.6\up{20.1} & \underline{44.9}\up{2.8} & \underline{34.7}\up{12.8} & \underline{50.9}\up{10.5} \\
         & AutoRefine & 50.5\up{3.2} & 82.8\up{32.3} & 35.6\dn{6.5} & 21.5\dn{0.4} & 47.6\up{7.2} & 47.1\dn{0.2} & 54.7\up{4.2} & 33.2\dn{8.9} & 22.5\up{0.6} & 39.4\dn{1.1} \\
         & Trace2Skill & 52.9\up{5.6} & 73.4\up{22.9} & 40.1\dn{2.0} & \underline{29.1}\up{7.2} & 48.9\up{8.4} & 52.9\up{5.6} & \underline{73.4}\up{22.9} & 40.1\dn{2.0} & 24.7\up{2.8} & 47.8\up{7.3} \\
         & Skill-Pro & 46.4\dn{1.0} & 50.7\up{0.2} & 32.1\dn{10.0} & 27.1\up{5.2} & 39.1\dn{1.4} & 43.2\dn{4.2} & 46.3\dn{4.2} & 32.7\dn{9.4} & 27.3\up{5.4} & 37.4\dn{3.1} \\
         & \textbf{SkillACE} & \textbf{69.7}\up{22.4} & \textbf{88.8}\up{38.3} & \textbf{54.9}\up{12.8} & \textbf{29.8}\up{7.9} & \textbf{60.8}\up{20.4} & \textbf{71.1}\up{23.8} & \textbf{93.5}\up{43.0} & \textbf{47.6}\up{5.5} & \textbf{38.3}\up{16.4} & \textbf{62.6}\up{22.2} \\
        \midrule
        \multirow{7}{*}{\shortstack[c]{Gemma-4-\\12B-it}} & No Skill & 58.6 & 49.3 & 41.4 & 32.0 & 45.3 & 58.6 & 49.3 & 41.4 & 32.0 & 45.3 \\
         & LLMGen & 52.2\dn{6.4} & \underline{86.8}\up{37.5} & 48.7\up{7.3} & 35.8\up{3.8} & 55.9\up{10.6} & 53.8\dn{4.8} & 74.6\up{25.3} & 49.3\up{7.9} & 33.8\up{1.8} & 52.9\up{7.6} \\
        \cmidrule(l){2-12}
         & SkillOpt & \underline{69.4}\up{10.8} & 83.8\up{34.5} & \underline{50.7}\up{9.3} & \textbf{39.0}\up{7.0} & \underline{60.7}\up{15.4} & 53.5\dn{5.1} & 81.6\up{32.3} & \underline{51.8}\up{10.4} & \textbf{41.0}\up{9.0} & 57.0\up{11.7} \\
         & AutoRefine & 61.9\up{3.3} & 82.6\up{33.3} & 46.1\up{4.7} & 35.7\up{3.7} & 56.6\up{11.3} & 54.3\dn{4.3} & 37.6\dn{11.7} & 46.0\up{4.6} & 35.1\up{3.1} & 43.3\dn{2.1} \\
         & Trace2Skill & 61.5\up{2.9} & 84.8\up{35.5} & 50.5\up{9.1} & \underline{37.1}\up{5.1} & 58.5\up{13.2} & \underline{61.5}\up{2.9} & \underline{84.8}\up{35.5} & 50.5\up{9.1} & 37.1\up{5.1} & \underline{58.5}\up{13.2} \\
         & Skill-Pro & 47.3\dn{11.3} & 48.0\dn{1.2} & 44.9\up{3.5} & 28.8\dn{3.2} & 42.2\dn{3.1} & 50.1\dn{8.5} & 53.5\up{4.2} & 44.3\up{2.9} & 29.1\dn{2.9} & 44.2\dn{1.1} \\
         & \textbf{SkillACE} & \textbf{72.3}\up{13.7} & \textbf{87.8}\up{38.5} & \textbf{56.2}\up{14.8} & 34.1\up{2.1} & \textbf{62.6}\up{17.3} & \textbf{69.2}\up{10.6} & \textbf{87.1}\up{37.8} & \textbf{60.4}\up{19.0} & \underline{38.1}\up{6.1} & \textbf{63.7}\up{18.4} \\
        \bottomrule
    \end{tabular}}
    \caption{Test results. Cross-model transfer reuses the Skills learned with
    the Qwen3.6-35B-A3B Solver. Subscripts give the change from No Skill with
    the same Solver. Best results are in \textbf{bold}, and second-best results
    are \underline{underlined}.}
    \label{tab:main-results}
\end{table*}

\subsection{Taxonomy-Guided Skill Organization and Updating}
\label{sec:skill-evolution}

Each error category $c$ corresponds to one fine-grained Skill $s_c$, which
holds the knowledge for correcting the error described by $d_c$, and the two
form a learning unit. The category--Skill correspondence is a mapping
$\phi_t: \mathcal C_t \to \mathcal S_t \cup \{\varnothing\}$, where categories
without a Skill map to $\varnothing$, and it is persistently stored throughout
evolution. When the Reflector merges one category into another, the Skill of
the merged category is reassigned to the remaining one.

Once the Reflector has categorized the failures, the Evolver selects the top
$K$ categories with the most failures and revises each corresponding Skill
using the failures $F_t^c$ of that category. It checks the existing Skill $s_c$
against the reflections and execution trajectories of these failures,
adding missing steps or revising inapplicable rules to produce a candidate
$s_c'$. If the category has no Skill yet, the Evolver creates a new one from
these failures as the candidate.

Each candidate first undergoes local validation, which compares it against the
batch's baseline library $\mathcal S_t$ on tasks in the target category and a
sample of successful tasks from the batch. The candidate must improve the
target category's mean score without decreasing the mean score on the
successful tasks. If it fails, the Evolver adjusts the revision based on
target tasks that remain unsolved and previously successful tasks whose scores
dropped, and regenerates a candidate again from $s_c$. This process repeats for
at most $R$ cycles.
Locally validated candidates then undergo global validation sequentially.
Starting from $\mathcal S_t$, each candidate is applied to the currently
accepted library and retained if the mean score on the validation set
$D_{\mathrm{val}}$ does not decrease. Algorithm~\ref{alg:skillace} summarizes
the full evolution procedure of SkillACE, where $\mathcal S \oplus s_c'$ denotes replacing or
adding the Skill of category $c$ with $s_c'$.

\section{Experiments}
\label{sec:experiments}

\subsection{Experimental Setup}
\label{sec:experimental-setup}

\paragraph{Benchmarks and metrics.} We evaluate SkillACE on four benchmarks:
ScienceWorld \citep{wang2022scienceworld} for interactive scientific experiments,
ALFWorld \citep{shridhar2021alfworld} for multi-step household tasks, WebShop
\citep{yao2022webshop} for product search and selection, and
SpreadsheetBench-Verified \citep{ma2024spreadsheetbench} for spreadsheet
manipulation. Each benchmark is split into evolution,
validation, and test tasks. Test tasks are never used during evolution, and the
split sizes are given in Table~\ref{tab:splits} (Appendix~\ref{sec:exp-details}).
We report the mean task score for ScienceWorld and the success rate for the other
benchmarks.

\paragraph{Baselines.} ReAct \citep{yao2023react} serves as the no-Skill
baseline. LLMGen uses a Skill generated directly by the Evolver
model without execution experience. We compare four methods that learn Skills from execution experience
with fixed model parameters. SkillOpt \citep{yang2026skillopt} and Trace2Skill
\citep{ni2026trace2skill} consolidate experience into a single Skill. The former
applies bounded edits gated by a validation set, and the latter hierarchically
merges patches proposed from individual trajectories. Skill-Pro
\citep{mi2026skillpro} and AutoRefine \citep{qiu2026autorefine} maintain
multiple Skills or multiple types of experience artifacts. The former revises
Skills with feedback aggregated per invoked Skill, and the latter compiles
experience into rules, Skills, or bounded subagents filtered by replay
validation. We do not take numbers from the original papers. Instead, we
reproduce every baseline from its official implementation under our setting,
with the same Solver, test sets, and interaction limits as SkillACE. Reproduction
details are given in Appendix~\ref{sec:baseline-details}.

\paragraph{Implementation details.} Qwen3.6-35B-A3B serves as the Solver for
evolution and the primary evaluation, and cross-model transfer uses Qwen3.5-9B
and Gemma-4-12B-it. The Reflector and Evolver use either GPT-5.5 \citep{openai2026gpt55} or
Qwen3.6-35B-A3B, and baselines learn Skills with the same model. Thinking is
disabled for the experiments in Table~\ref{tab:main-results}. Each task allows
at most 60 steps on ALFWorld, 15 steps on WebShop, 100 steps on ScienceWorld,
and 30 interaction turns on SpreadsheetBench. SkillACE processes 512 evolution
tasks per batch. Because the SpreadsheetBench evolution set is small, each batch there
uses all 120 tasks. Each batch selects at most $K=3$ error categories, with up
to $R=3$ candidate-generation cycles per category. The Skill library is fixed at
test time, and test results average three complete executions on all benchmarks.
The number of evolution epochs and the initial Skills for
each benchmark are given in Table~\ref{tab:hparams} (Appendix~\ref{sec:impl-details}).

\subsection{Main Results}
\label{sec:main-results}

\noindent\textbf{SkillACE outperforms all baselines on every benchmark.} With
Qwen3.6-35B-A3B serving as the Solver, Reflector, and Evolver, SkillACE
achieves the best result on all four benchmarks and raises the average score
from 61.8 to 73.3 (Table~\ref{tab:main-results}). Its largest margins occur on
ScienceWorld and SpreadsheetBench, where it reaches 78.9 and 74.5\%, compared
with 66.3 and 68.3\% for the second-best method, SkillOpt. Baseline gains are
less consistent. Trace2Skill leaves the average score nearly unchanged at 61.7
and falls below No Skill on SpreadsheetBench. AutoRefine and Skill-Pro improve
only WebShop and drop on the other three benchmarks, lowering the average score
to 58.7 and 58.4.

\noindent\textbf{A stronger Reflector and Evolver further raises the gains.} With
GPT-5.5 as the Reflector and Evolver, SkillACE again achieves the best result
on all four benchmarks and reaches an average score of 76.4. Its lead over the
baselines remains, and AutoRefine and Skill-Pro still fall below No Skill on
average. This suggests that organizing experience by error contributes to the gains
beyond the capability of the model that writes the Skills.

\noindent\textbf{SkillACE Skills generalize within and across model families.}
Cross-model transfer tests whether the Skills carry knowledge beyond the Solver
they were learned with. Without re-evolution, the Skills learned with
Qwen3.6-35B-A3B in all three roles improve both unseen Solvers. On Qwen3.5-9B
from the same family, SkillACE achieves the best result on all four benchmarks
and raises ALFWorld success from 50.5\% to 93.5\%. On Gemma-4-12B-it from a
different family, it achieves the best average score and raises ALFWorld
success from 49.3\% to 87.1\%, compared with 84.8\% for Trace2Skill. Skills
learned with GPT-5.5 transfer similarly.

\section{Analysis}
\label{sec:analysis}

\subsection{RQ1: Does an Adaptive Error Taxonomy Outperform Predefined Ones?}
\label{sec:adaptive-taxonomy}

The adaptive taxonomy of SkillACE induces error categories from the failures
observed during evolution and keeps expanding as new errors emerge. To verify
the advantage of our adaptive taxonomy, we keep all other components of SkillACE fixed on
SpreadsheetBench and replace the adaptive taxonomy with three taxonomies that
are specified before evolution and remain unchanged afterwards. The first is
the generic agent error taxonomy proposed by AgentDebug
\citep{zhu2025agentdebug}. The second is enumerated by GPT-5.5 from the task
descriptions alone. The third is induced from failure trajectories by AdaMAST
\citep{cemri2026adamast} (Appendix~\ref{sec:taxonomy-details}).
Besides the test
success rate, we count the candidate Skill revisions and those passing local
and global validation. The ratio of global to local passes (G/L) measures how
often a revision that repairs its target category also leaves other tasks
intact. As shown in Table~\ref{tab:adaptive-taxonomy}, our adaptive taxonomy
outperforms all three.

\begin{table}[t]
    \centering
    \fontsize{8}{10}\selectfont
    \setlength{\tabcolsep}{2pt}
    \begin{tabular*}{\columnwidth}{@{\extracolsep{\fill}}lr@{\extracolsep{0pt}}l@{\extracolsep{\fill}}rrrr@{}}
        \toprule
        \multirow{2}{*}{Taxonomy} & \multicolumn{2}{c}{\multirow{2}{*}{SR (\%)}} & \multicolumn{4}{c}{Candidate validation} \\
        \cmidrule(l){4-7}
        & \multicolumn{2}{c}{} & Cand. & Local & Global & G/L (\%) \\
        \midrule
        No Skill & 61.55 & \phantom{\up{00.00}} & -- & -- & -- & -- \\
        Generic & 72.74 & \up{11.19} & 29 & 16 & 3 & 18.75 \\
        AdaMAST & 76.55 & \up{15.00} & 36 & 12 & 5 & 41.67 \\
        Prior-enumerated & 79.40 & \up{17.86} & 35 & 13 & 5 & 38.46 \\
        Adaptive (ours) & 82.26 & \up{20.71} & 35 & 13 & 7 & 53.85 \\
        \bottomrule
    \end{tabular*}
    \caption{Taxonomy comparison on SpreadsheetBench. SR denotes success rate.
    Cand., Local, and Global count candidate Skill revisions and those passing
    local and global validation, and G/L is Global divided by Local.}
    \label{tab:adaptive-taxonomy}
\end{table}

\begin{figure}[t]
    \centering
    \includegraphics[width=\columnwidth]{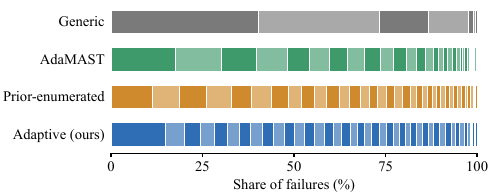}
    \caption{Distribution of evolution failures over error categories on
    SpreadsheetBench. Each bar is one taxonomy, with categories sorted by size.}
    \label{fig:taxonomy-distribution}
\end{figure}

\noindent\textbf{Generic categories are too coarse and mix errors with different
causes, so their revisions tend to interfere with other tasks.} The generic
taxonomy divides errors by agent module, whereas the causes of spreadsheet
errors do not align with modules, so many failures with different causes fall
into the same category. As shown in Figure~\ref{fig:taxonomy-distribution},
the three largest generic categories absorb 86.7\% of the failures, compared
with 24.4\% for the adaptive taxonomy. A revision written for such a mixed
category can repair part of its failures while changing behavior that other
tasks rely on, so the generic taxonomy has the lowest G/L ratio in
Table~\ref{tab:adaptive-taxonomy}.

\noindent\textbf{A prior-enumerated taxonomy cannot anticipate errors that arise only
during execution.} The prior-enumerated categories are derived from task
descriptions alone, so they describe what a task asks for and miss how the
agent fails while executing it. For example, the agent may analyze the sheet
structure and then end without writing the target cells. The adaptive
taxonomy forms dedicated categories for such errors from trajectories, whereas
the prior-enumerated taxonomy has no counterpart and places these failures in
semantically related categories such as conditional operator precedence. The
revisions for those categories then target the wrong cause and tend to
interfere with other tasks.

\noindent\textbf{AdaMAST induces categories from failures, but a fixed scheme
constrains them, so they also mix errors with different causes.} AdaMAST
places all its categories under three fixed axes, system, role, and domain, which divide failures by the
part of the agent system where the fix belongs. Spreadsheet tasks are solved by
a single agent, and most of their errors call for task knowledge, so most
failures fall on the domain axis and concentrate in a few broad categories.
As shown in Figure~\ref{fig:taxonomy-distribution}, the three largest AdaMAST
categories hold 39.7\% of the failures, compared with 24.4\% for the adaptive
taxonomy with a similar number of categories. These broad categories mix
failures that need different corrections.

\subsection{RQ2: Does Organizing and Updating Skills by Error Category Improve Performance?}
\label{sec:skill-granularity}

We compare SkillACE with a global variant that maintains a single Skill and a
scenario variant that maintains one Skill per task scenario. The three share the same taxonomy and validation
and differ only in which Skill a revision is written to. As shown in
Table~\ref{tab:skill-granularity}, SkillACE outperforms both variants.
Scenario-level Skills do not outperform a single global Skill, so finer-grained
Skills bring no benefit by themselves. The gain comes from organizing Skills
around errors instead of tasks.

\begin{table}[t]
    \centering
    \fontsize{9}{10}\selectfont
    \setlength{\tabcolsep}{1.5pt}
    \begin{tabular*}{\columnwidth}{@{\extracolsep{\fill}}lr@{\extracolsep{0pt}}l@{\extracolsep{\fill}}rrrr@{}}
        \toprule
        Variant & \multicolumn{2}{c}{SR (\%)} & Cand. & Local & Global & G/L (\%) \\
        \midrule
        No Skill & 61.55 & \phantom{\up{00.00}} & -- & -- & -- & -- \\
        Global & 75.48 & \up{13.93} & 42 & 10 & 3 & 30.00 \\
        Scenario & 74.05 & \up{12.50} & 38 & 9 & 3 & 33.33 \\
        Mixed failures & 77.86 & \up{16.31} & 35 & 12 & 4 & 33.33 \\
        No persist. & 77.02 & \up{15.48} & 38 & 12 & 4 & 33.33 \\
        \midrule
        Ours & \textbf{82.26} & \up{20.71} & 35 & \textbf{13} & \textbf{7} & 53.85 \\
        \bottomrule
    \end{tabular*}
    \caption{Skill organization and updating on SpreadsheetBench. Columns
    follow Table~\ref{tab:adaptive-taxonomy}.}
    \label{tab:skill-granularity}
\end{table}

\begin{figure}[t]
    \centering
    \includegraphics[width=0.49\columnwidth]{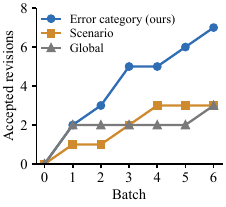}\hfill
    \includegraphics[width=0.49\columnwidth]{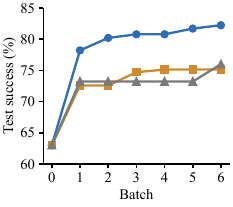}
    \caption{SpreadsheetBench evolution: cumulative accepted revisions (left)
    and test success (right). Batch 0 uses no Skills.}
    \label{fig:skill-organization-batches}
\end{figure}

\noindent\textbf{Error-specific revisions interfere with other tasks less often than
revisions to shared Skills.} Revisions written to a shared Skill are less
likely to pass local validation, and those that pass are less likely to pass
global validation (Table~\ref{tab:skill-granularity}). A shared Skill serves all tasks, so a change
aimed at one class of failures easily alters behavior that other tasks rely on,
whereas error-specific organization confines each change to the Skill of its
error. As a result, a shared Skill soon stops improving. As shown in
Figure~\ref{fig:skill-organization-batches}, the global Skill accepts no
revision from the second to the fifth batch and its test success stays flat,
whereas SkillACE accepts revisions in five of the six batches and rises
steadily.

\noindent\textbf{Failures from other categories weaken the revision of a Skill.}
We retain the error taxonomy and category-specific Skill library but replace
the failures used for revision. For each selected category, the mixed-failure
variant draws the same number of failures uniformly at random from the full
batch, instead of using only failures of that category. The category
description, existing Skill, and validation procedure remain unchanged.
This variant falls 4.4 points below SkillACE
(Table~\ref{tab:skill-granularity}), which supports using failures of the same
category to revise its corresponding Skill.

\subsection{RQ3: Do Later Failures of a Category Keep Improving the Same Skill?}
\label{sec:repeated-revision}

SkillACE keeps the correspondence between categories and Skills across
batches, so later failures of a category continue to revise its existing Skill,
and experience accumulates in that Skill. To verify the benefit of this
persistent correspondence, we compare SkillACE with a variant that creates a
new Skill for each selected category instead of revising the existing one, and
we examine whether the same Skill keeps improving across successive revisions.

\noindent\textbf{Creating new Skills instead of revising existing ones stalls
evolution.} The variant (No persist.\ in Table~\ref{tab:skill-granularity})
keeps the taxonomy and validation of SkillACE but, similar to Ratchet
\citep{zhang2026ratchet}, creates a new Skill for each selected category in
every batch and never revises existing Skills. Its success rate drops from
82.26\% to 77.02\%, and no new Skill is accepted after the third batch. By then
every selected category already has a Skill, and the variant writes a second
one from the current failures without seeing the first. The two Skills overlap
or conflict, so none of the new Skills passes global validation.

\begin{figure}[t]
    \centering
    \includegraphics[width=\columnwidth]{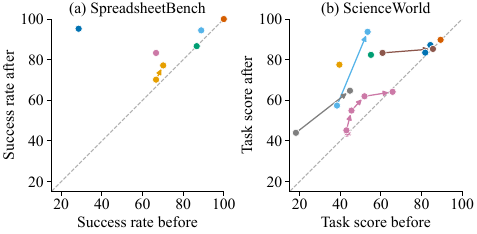}
    \caption{Performance before and after each accepted Skill update on the
    failed tasks of its category. Each color denotes one error category, arrows
    connect successive updates of the same category, and points above the
    diagonal indicate improvement.}
    \label{fig:revision-pairs}
\end{figure}

\noindent\textbf{The same Skill keeps improving with later failures of its category.}
For each accepted update on SpreadsheetBench and ScienceWorld, we evaluate the
Skill libraries before and after the update, which differ only in the target
Skill, on all tasks assigned to that category during evolution. In
Figure~\ref{fig:revision-pairs}, each point is one accepted Skill update. Each
color denotes one error category, and arrows connect successive updates of the Skill of the
same category. Most points lie above the diagonal, and the chains move toward
the upper right, so later revisions build on earlier versions to further
improve performance.
For example, the genetic-dominance tasks in ScienceWorld require the agent to
grow offspring from parent plants and infer whether a trait is dominant. The
category of choosing a dominance answer without growing offspring receives four
accepted updates, which raise the task score from 43.53 to 61.98. The first
version only asks the agent to observe contrasting traits, so the agent still
judges from the parent plants. Later versions require inferring dominance from
offspring traits, keeping parental traits apart from offspring evidence, and
confirming that planting succeeded before waiting for growth. Each revision
fixes a new gap of the same error, and all of them accumulate in one Skill.

\subsection{RQ4: Does SkillACE Cost More Context?}
\label{sec:context-cost}

We measure the Skill text that each test task brings into
the Solver's context, counting the Skill catalog and every Skill body the
Solver reads, and compare SkillACE with the two single-Skill baselines using
the libraries evaluated in Table~\ref{tab:main-results}.
Appendix~\ref{sec:skill-token-footprint} gives the counting details.

\begin{table}[t]
    \centering
    \small
    \setlength{\tabcolsep}{2pt}
    \begin{tabular*}{\columnwidth}{@{\extracolsep{\fill}}lcccc@{}}
        \toprule
        & \multicolumn{3}{c}{Per task} & \multicolumn{1}{c}{Library} \\
        \cmidrule(lr){2-4} \cmidrule(l){5-5}
        Benchmark & SkillOpt & Trace2Skill & SkillACE & SkillACE \\
        \midrule
        ScienceWorld & 6.8 & 2.7 & \textbf{2.0} & 8.6 \\
        ALFWorld & 2.8 & \textbf{2.2} & 3.0 & 9.5 \\
        Spreadsheet & 3.4 & 3.4 & \textbf{3.2} & 5.0 \\
        WebShop & 6.5 & 2.6 & \textbf{1.6} & 1.6 \\
        \midrule
        Avg & 4.9 & 2.7 & \textbf{2.5} & 6.2 \\
        \bottomrule
    \end{tabular*}
    \caption{Skill tokens (thousands) brought into the Solver's context per
    test task, and the total size of the final SkillACE library. The lowest
    per-task value on each benchmark is in bold.}
    \label{tab:context-cost}
\end{table}

\noindent\textbf{The agent effectively loads only the relevant Skills and thus
uses less context than single-Skill methods.} As shown in
Table~\ref{tab:context-cost}, SkillACE brings the least Skill text into the
Solver's context on average. The Solver loads only the fine-grained Skills
relevant to the current task, whereas a single-Skill baseline places its
entire Skill in every task. Combined with the results in
Table~\ref{tab:main-results}, SkillACE loads fewer Skill tokens yet achieves
higher performance, showing that it uses Skills more efficiently than
single-Skill methods.

\section{Conclusion}
\label{sec:conclusion}

We presented SkillACE, a framework that enables LLM agents to accumulate
experience in Skills organized by the errors they make. By building an error
taxonomy from scratch, SkillACE groups failures by their underlying errors
instead of by task, so each revision addresses a single error.
To make these corrections cumulative, we pair each error category with one
fine-grained Skill and keep the pairing throughout evolution, so recurring
failures keep refining the same Skill. Finally, local and global validation
admit a revision only when it repairs its target category without degrading
other tasks. Results on four agentic benchmarks confirm that SkillACE
outperforms existing Skill evolution methods and that its Skills transfer
across LLMs. Our findings validate that errors are an
effective unit for accumulating agent experience.

\section*{Limitations}

\paragraph{SkillACE requires evaluable task feedback.} It uses this feedback to
identify failures and determine whether Skill revisions are effective. The benchmarks in this study provide automatic
scores. For open-ended tasks with subjective outcomes or delayed feedback,
further work is needed to obtain feedback that supports error categorization
and reliable acceptance of Skill revisions.

\paragraph{Our evaluation only covers text-based environments.} The four benchmarks
present observations and actions as text. Agents that act on visual observations, such as GUI or embodied
agents with image inputs, may exhibit errors that are hard to describe from
textual trajectories alone. Extending error categorization and Skill evolution
to such multimodal settings is a direction for future work.

\bibliography{custom}

\clearpage
\appendix
\section{Experimental Setup}
\label{sec:exp-details}

\paragraph{Benchmarks.} The four benchmarks cover interactive environments with
different action spaces and feedback, and each contains many tasks that share
recurring error patterns, which makes them suitable for accumulating experience
across tasks.
\begin{itemize}
    \item \textbf{ScienceWorld} \citep{wang2022scienceworld} is a text-based
    environment for elementary-school science experiments. Its 30 task types
    span ten topics, such as changes of state, electrical circuits, and
    Mendelian genetics. The agent moves between rooms and manipulates objects
    and instruments through text commands, and each task returns a score from
    0 to 100 that credits partial progress. Tasks are long and require
    procedural knowledge about the experiment, such as which evidence suffices
    to determine a genetic trait.
    \item \textbf{ALFWorld} \citep{shridhar2021alfworld} is a text-based
    household environment aligned with the ALFRED embodied benchmark. Its six
    task types ask the agent to find, place, examine, clean, heat, or cool
    objects, and success is binary. The \texttt{valid\_unseen} split uses rooms
    and object layouts absent from training, so it measures whether learned
    experience transfers beyond the evolution tasks.
    \item \textbf{WebShop} \citep{yao2022webshop} simulates an e-commerce
    website with over a million real products. Given an instruction describing
    a desired product and its attributes, the agent searches, browses result
    pages, selects options, and buys an item. The reward measures how well the
    purchased item matches the requested attributes, and a task succeeds only
    when all of them match.
    \item \textbf{SpreadsheetBench-Verified} \citep{ma2024spreadsheetbench}
    contains real spreadsheet questions collected from online Excel forums,
    with human-verified answers. Each task provides a workbook and an
    instruction, and the agent edits the workbook by executing Python code.
    Tasks are either cell-level, writing results into specified cells, or
    sheet-level, modifying the sheet as a whole. The output is graded by
    comparing cell values with a reference workbook, so the agent must produce
    verifiable values rather than plausible-looking formulas.
\end{itemize}

\paragraph{Splits and metrics.} Table~\ref{tab:splits} lists the splits.
ScienceWorld reports the mean task score on a 0--100 scale; its test set is a
fixed sample of 1,000 of the
1,819 test tasks. ALFWorld uses the full \texttt{valid\_unseen}
split as the test set, and its validation set combines 481 held-out training
tasks with the 140 \texttt{valid\_seen} tasks. SpreadsheetBench-Verified
contains 400 tasks, so the 120 non-test tasks serve both as the evolution set
and as the validation set for global validation, and the remaining 280 tasks
form the test set. A SpreadsheetBench task succeeds when every answer region of
the output workbook matches the reference workbook after recalculation.

\begin{table}[h]
    \centering
    \small
    \begin{tabular}{@{}lrrr@{}}
        \toprule
        Benchmark & Evolve & Valid & Test \\
        \midrule
        ScienceWorld & 3,592 & 768 & 1,000 \\
        ALFWorld & 3,072 & 621 & 134 \\
        WebShop & 4,096 & 512 & 500 \\
        SpreadsheetBench & 120 & 120$^\dagger$ & 280 \\
        \bottomrule
    \end{tabular}
    \caption{Data splits. $^\dagger$The evolution tasks also serve as the
    validation set.}
    \label{tab:splits}
\end{table}

\paragraph{Interaction limits and decoding.} Each episode allows at most 100
environment steps on ScienceWorld, 60 on ALFWorld, 15 on WebShop, and 30
interaction turns on SpreadsheetBench. The Solver decodes with temperature 1.0,
top-$p$ 0.95, and top-$k$ 20, and generates at most 2,048 tokens per turn. Thinking is disabled for every model in every
stage, including evolution, validation, and testing.

\paragraph{Models and serving.} Qwen3.6-35B-A3B serves as the Solver for
evolution and primary evaluation, and Qwen3.5-9B and Gemma-4-12B-it serve as
Solvers in cross-model transfer. The Reflector and Evolver use either GPT-5.5
(\texttt{gpt-5.5-2026-04-24}) through its API or Qwen3.6-35B-A3B. Open-weight
models are served with vLLM, using four replicas with tensor parallelism 2 on
each 8-GPU node and a maximum context length of 262,144 tokens.

\section{Baseline Reproduction}
\label{sec:baseline-details}

We reproduce every baseline from its official repository at a fixed commit
(Table~\ref{tab:baseline-sources}) and connect it to our environments through
an adapter that only provides task execution and feedback, leaving the
learning procedure of each method unchanged. All baselines use the same Solver,
test sets, interaction limits, and decoding settings as SkillACE, and they
learn Skills with the same model as the SkillACE Reflector and Evolver in each
setting.

\begin{table}[h]
    \centering
    \footnotesize
    \setlength{\tabcolsep}{3pt}
    \begin{tabular}{@{}lll@{}}
        \toprule
        Method & GitHub repository & Commit \\
        \midrule
        SkillOpt & \texttt{microsoft/SkillOpt} & \texttt{79124b3} \\
        AutoRefine & \texttt{AutoRefine/Autorefine} & \texttt{58ebf90} \\
        Trace2Skill & \texttt{Qwen-Applications/Trace2Skill} & \texttt{3d0b52a} \\
        Skill-Pro & \texttt{Miracle1207/Skill-Pro} & \texttt{3be7a9b} \\
        \bottomrule
    \end{tabular}
    \caption{Official implementations used for the baselines.}
    \label{tab:baseline-sources}
\end{table}

\paragraph{Evolution schedule.}
The methods follow two kinds of
learning procedures. SkillOpt and Skill-Pro iterate over the evolution set in
epochs of fixed-size batches, as SkillACE does. SkillOpt selects its final
Skill on the validation set, and Skill-Pro accepts a candidate only when its
estimated improvement exceeds 0.001. AutoRefine and Trace2Skill instead learn
from a fixed pool of trajectories. Following their original protocols, the pool
contains a stratified sample of 200 evolution tasks on ScienceWorld and
ALFWorld, all 120 evolution tasks on SpreadsheetBench, and all 4,096 evolution
tasks on WebShop. AutoRefine revisits
this pool for several epochs, whereas Trace2Skill collects trajectories once,
analyzes them, and merges the resulting patches into a single Skill.

\section{Implementation Details of SkillACE}
\label{sec:impl-details}

\paragraph{Skills and progressive loading.} Each Skill is a versioned Markdown
document with a name, a short description, and an applicability line that
names the task scenario it applies to. At the start of a task, the Solver sees
a catalog that lists the name, description, applicability, and section
headings of every active Skill, but not the Skill bodies. It loads the full
body of a Skill through a \texttt{read\_skill} tool and is instructed to read
every Skill whose applicability matches the current task.

\paragraph{Reflector.} For each failed task, the Reflector receives the task
instruction, the Solver's final output and reasoning, a projection of the
execution trace, and the outcome feedback. It writes a reflection on the
erroneous behavior, assigns the failure to a task scenario, and either assigns
it to an existing error category or proposes a new one. To maintain the
taxonomy, it can add, edit, merge, and delete both task scenarios and error
categories; category descriptions and the content of their Skills are visible
during assignment.

\paragraph{Evolver.} Among categories with at least two failures in the batch,
the Evolver selects the $K=3$ categories with the most failures. For each
category, it receives the failures, their reflections and trajectories, and the
existing Skill of the category, and it either edits that Skill or creates a new
one. When the Reflector merges a category into another, the Skill of the merged
category is migrated to the remaining category, and its applicability line is
updated accordingly; when a category is deleted, its Skill is retired.

\paragraph{Local validation.} The candidate and the baseline library are each
evaluated on tasks of the target category, with scores normalized to $[0,1]$.
A candidate passes if two conditions hold. First, its mean score on the target
tasks exceeds that of the baseline library by at least a margin $\delta$, a
hyperparameter set to 0.05 in all experiments. Second, the Solver
loads the revised Skill on the target tasks, and the score on these tasks
improves on average, so the gain can be attributed to the revision. A failed
candidate is regenerated from the same parent Skill with feedback on target
tasks that remain unsolved, for at most $R=3$ cycles.

\paragraph{Global validation.} Locally validated candidates are applied one at
a time, in the order of their categories, to the currently accepted library. A
candidate is retained if the mean score on the validation set does not
decrease.

\paragraph{Per-benchmark settings.} Table~\ref{tab:hparams} summarizes the
evolution settings of SkillACE.

\begin{table}[h]
    \centering
    \small
    \setlength{\tabcolsep}{3pt}
    \begin{tabular}{@{}lrrrl@{}}
        \toprule
        Benchmark & Batch & Epochs & Scen. & Init.\ Skills \\
        \midrule
        ScienceWorld & 512 & 3 & 10 & None \\
        ALFWorld & 512 & 4 & 6 & None \\
        WebShop & 512 & 4 & 5 & None \\
        SpreadsheetBench & 120 & 6 & 2 & None \\
        \bottomrule
    \end{tabular}
    \caption{SkillACE evolution settings. Scen. is the number of initial task
    scenarios. All benchmarks use $K=3$ and $R=3$. }
    \label{tab:hparams}
\end{table}

\section{Context Cost of Skills}
\label{sec:skill-token-footprint}

Table~\ref{tab:context-cost} in Section~\ref{sec:context-cost} uses the final
Skill libraries evaluated in Table~\ref{tab:main-results}. For each test task,
we count the Skill catalog in the system prompt and every Skill body the Solver
reads, and average over test tasks. The library size counts the text of all
active SkillACE Skills. Tokens are counted with the Qwen3.6-35B-A3B tokenizer.

\section{Stability Across Evolution Task Orders}
\label{sec:order-seeds}

SkillACE evolves its Skill library batch by batch. To measure how much the
final library depends on the order of evolution tasks, we evolve SkillACE on
SpreadsheetBench under three task orders, given by data-order seeds 42, 43, and
44, with all other randomness fixed at seed 42. Qwen3.6-35B-A3B serves as the
Solver, Reflector, and Evolver. Each run starts from an empty library and
evolves for six epochs over the 120 evolution tasks, and its final library is
evaluated three times on the 280 test tasks. The run with seed 42 is the one
reported in Table~\ref{tab:main-results}.

\begin{table}[t]
    \centering
    \small
    \setlength{\tabcolsep}{6pt}
    \begin{tabular}{@{}lrr@{}}
        \toprule
        Task order & Evolve (\%) & Test (\%) \\
        \midrule
        Seed 42 & 76.39 & 74.52 \\
        Seed 43 & 78.06 & 76.43 \\
        Seed 44 & 77.22 & 77.62 \\
        \midrule
        Mean $\pm$ SD & $77.22 \pm 0.83$ & $76.19 \pm 1.56$ \\
        \bottomrule
    \end{tabular}
    \caption{SkillACE on SpreadsheetBench under three evolution task orders.
    Evolve is the success rate on the evolution tasks, averaged over the three
    evaluations at the last accepted global validation. SD is the sample
    standard deviation.}
    \label{tab:order-seeds}
\end{table}

As shown in Table~\ref{tab:order-seeds}, test success stays within a 3.1-point
range across the three orders, with a mean of 76.19\% and a standard deviation
of 1.56 points. Every order exceeds SkillOpt, the strongest baseline in the same
all-Qwen setting at 68.3\%, by more than 6 points.

\section{Taxonomy Comparison Details}
\label{sec:taxonomy-details}

This appendix specifies the three predefined taxonomies compared in
Section~\ref{sec:adaptive-taxonomy} and reports how failures distribute over
their categories. Each variant keeps the other components of SkillACE fixed,
and the predefined taxonomies stay unchanged throughout evolution.

\paragraph{Construction.} SpreadsheetBench divides its tasks into cell-level
manipulation and sheet-level manipulation, and every taxonomy except AdaMAST
uses these two types as task scenarios.
\begin{itemize}
    \item \textbf{Generic.} The five module-level error types of
    AgentErrorTaxonomy from AgentDebug \citep{zhu2025agentdebug}, namely
    memory, reflection, planning, action, and system errors, are placed under
    each scenario, giving 10 categories.
    \item \textbf{Prior-enumerated.} GPT-5.5 enumerates categories from the raw
    descriptions and scenario labels of the 120 evolution tasks, without seeing
    any trajectory or outcome. We set the number of categories to 43, the size
    of the final adaptive taxonomy, so that the two taxonomies differ only in
    whether their categories are derived from failures.
    \item \textbf{AdaMAST.} AdaMAST \citep{cemri2026adamast} induces 41
    categories from the failures of the first batch, which runs without Skills
    and is also the first batch that the SkillACE Reflector categorizes. The
    categories are organized along the system, role, and domain axes and are
    shared by all tasks.
\end{itemize}

\paragraph{Generic categories.} The cell-level action category receives
failures from 59 tasks. Of these, 47 also fail under the adaptive taxonomy,
which assigns them to 26 different categories, ranging from misreading
conditional aggregation rules and writing formulas instead of computed values
to ending without submission after partial edits. Module boundaries also split
errors with the same cause: misreading conditional aggregation rules appears
under both the action and the planning categories.

\paragraph{Prior-enumerated categories.} The eight execution-process categories
of the adaptive taxonomy cover failures on 17 tasks. The prior-enumerated run
places these failures in semantic categories such as conditional operator
precedence, while six of its 43 categories never receive a failure during the
six epochs.

\paragraph{AdaMAST categories.} Eight of the 11 system-level categories, such
as misreading tool responses and losing state between steps, never receive a
failure during the six epochs. The largest category, conditional criteria logic
errors, holds 17.5\% of all failures from 23 tasks.

\section{Growth of the Error Taxonomy}
\label{sec:taxonomy-growth}

\begin{figure*}[t]
    \centering
    \includegraphics[width=0.76\textwidth]{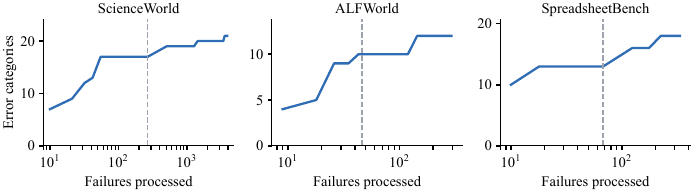}
    \caption{Number of error categories against the cumulative number of
    failures processed during evolution, on a logarithmic scale. The dashed
    line marks the end of the first batch.}
    \label{fig:taxonomy-growth}
\end{figure*}

Figure~\ref{fig:taxonomy-growth} tracks the number of error categories as the
Reflector processes failures, in the evolution runs of
Table~\ref{tab:main-results} with Qwen3.6-35B-A3B as the Reflector and
Evolver. The Reflector categorizes the failures of the first batch in
sequential chunks, so the curve follows the taxonomy within that batch;
afterwards, it is sampled at the end of each later batch. Categories are
created quickly at the start, and later batches add new categories at a
decreasing rate as new errors appear, reaching 12 to 21 categories per
benchmark.

\section{Case Study: Rule Interference in a Single Shared Skill}
\label{sec:case-study}

\begin{figure*}[t]
    \centering
    \includegraphics[width=\textwidth]{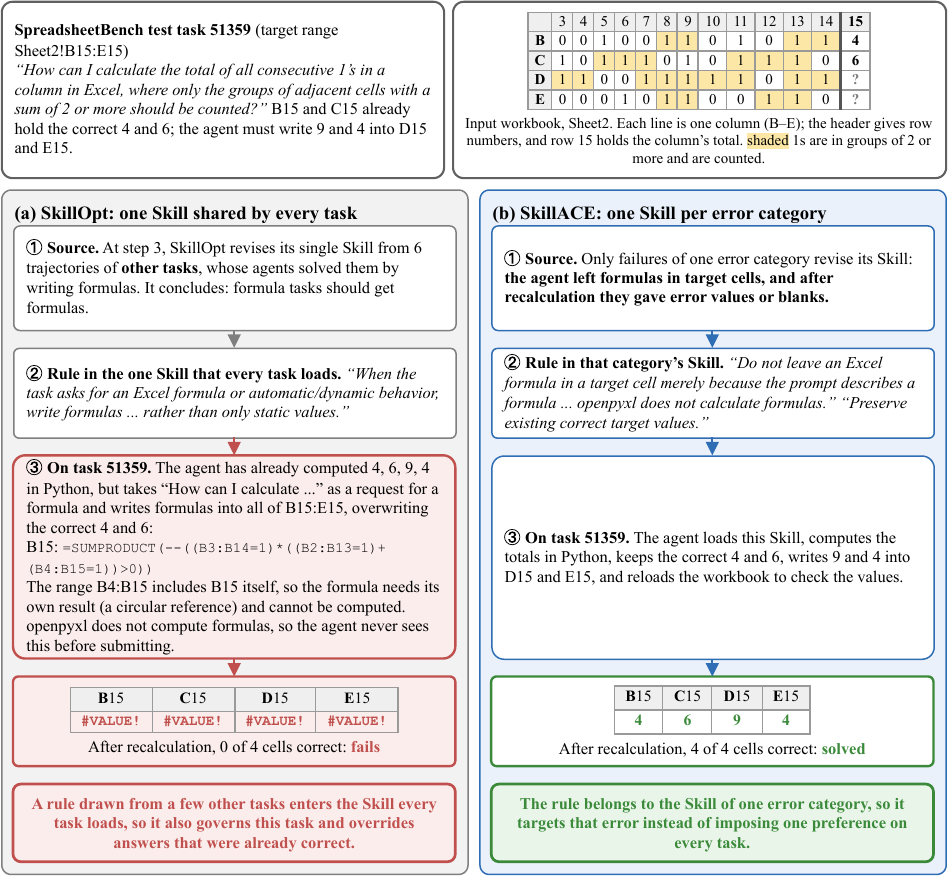}
    \caption{SpreadsheetBench test task 51359 under the final Skills of SkillOpt
    and SkillACE. (a) SkillOpt keeps one Skill for every task, so a formula rule
    drawn from six trajectories of other tasks also applies here and leads the
    agent to overwrite correct values with formulas that evaluate to
    \texttt{\#VALUE!}. (b) In SkillACE, the relevant rule sits in the Skill of
    one error category, is revised only from that category's failures, and
    targets that error. Quotations are verbatim from the evolution records,
    the Skills, and the agent's code comments.}
    \label{fig:case-study}
\end{figure*}

Figure~\ref{fig:case-study} follows one SpreadsheetBench test task under the
final Skills of SkillOpt and SkillACE. Task 51359 asks: ``How can I calculate
the total of all consecutive 1's in a column in Excel, where only the groups of
adjacent cells with a sum of 2 or more should be counted?'' Columns B to E of
Sheet2 hold 0s and 1s in rows 3 to 14, and the answers go in B15:E15. The input
workbook already contains the correct totals 4 and 6 in B15 and C15, so the
agent only needs to write 9 and 4 into D15 and E15. Without a Skill, the agent
computes the totals in Python and writes them into the cells as plain numbers,
solving the task in five of six runs. With SkillOpt's Skill, however, the
agent follows a rule favoring formulas and replaces correctly computed
answers with incorrect formulas, causing the task to fail.

\paragraph{SkillOpt: a rule from other tasks reaches every task.} At step 3
of evolution, SkillOpt reviews a minibatch of six trajectories from other tasks,
observes that ``for formula tasks, successful agents chose formulas that
preserve dynamic workbook behavior rather than only writing currently correct
static values,'' and appends the rule ``when the task asks for an Excel formula
or automatic/dynamic behavior, write formulas in the target cells rather than
only static values, unless the instruction explicitly asks for final values.''
The revision raises the selection score and is accepted, and a reworded version
of the rule remains in the final Skill. Because SkillOpt keeps one Skill for the whole
domain, the rule applies to every task. On task 51359, the agent first computes
the correct totals 4, 6, 9, and 4 in Python, then reasons that ``the
instruction says `How can I calculate...' - this is asking for a formula'' and
writes a \texttt{SUMPRODUCT} formula into all four target cells, overwriting the
two prefilled answers. The range B4:B15 contains B15 itself, so the formula
refers to its own cell, and after recalculation every target cell shows
\texttt{\#VALUE!}. Two of the three runs fail in this way, with none of the
four target cells correct.
The grader recalculates the formulas in LibreOffice and compares the resulting
cell values with the reference answers. The agent cannot detect these formula
errors by reopening the workbook with openpyxl, because openpyxl saves
formulas without evaluating them.

\paragraph{SkillACE: a rule in one category's Skill targets that error.} The
relevant knowledge sits in the Skill of an error category whose failures left
formulas in target cells that evaluated to errors or blanks after
recalculation. Only failures of this category revise the Skill, so its rule
addresses that error directly: ``Do not leave an Excel formula in a target
cell merely because the prompt describes a formula,'' because openpyxl writes
formulas without calculating them, and ``preserve existing correct target
values \dots; do not overwrite them with formulas.'' The agent loads this Skill
in all three runs. It computes the four totals in Python, notes that ``B15 and
C15 are already correct (4 and 6),'' writes 9 and 4 into D15 and E15, and checks
the reloaded target range. All three runs succeed.

The two outcomes follow from how each method organizes its Skills. SkillOpt
writes every rule into one Skill that all tasks load, so a rule drawn from a
few tasks also governs tasks it was never derived from and overrides behavior
that was already correct. SkillACE assigns each rule to the Skill of one error
category and revises it only from that category's failures, so the rule
corrects that error without imposing a single preference on every task.

\section{Learned Taxonomy and Skill Examples}
\label{sec:learned-skills}

We show three error categories from a ScienceWorld taxonomy learned by SkillACE with one Skill per error category. Table~\ref{tab:top-categories} summarizes the categories and their corrections, and Figures~\ref{skill-example-friction}--\ref{skill-example-classification} give the complete text of each Skill.

\begin{table*}[t]
\centering
\small
\begin{tabular}{@{}p{0.17\textwidth}p{0.43\textwidth}p{0.34\textwidth}@{}}
\toprule
Task scenario & Error category & Skill correction \\
\midrule
Friction & Fails to complete timed inclined-plane comparison and focus on the selected surface before step limit & Run bounded trials with the same object on both planes, then focus on the selected plane. \\
Life spans & Chooses animal by visible life stage instead of comparing species life spans & Compare species lifespans rather than the visible age or egg stage. \\
Object classification & Focused on an object that does not satisfy the requested living/plant/animal/non-living category & Prove that an object belongs to the requested category before focusing on it. \\
\bottomrule
\end{tabular}
\caption{Three ScienceWorld error categories and the corrections given by their Skills.}
\label{tab:top-categories}
\end{table*}

\begin{figure*}[p]
\begin{skillbox}[skillfloat]{Example: Friction Skill}
\textbf{Skill Definition: Friction: bounded inclined-plane trials before final focus}
\begin{itemize}
\item \textbf{Error category}: Fails to complete timed inclined-plane comparison and focus on the selected surface before step limit
\item \textbf{Applicability}: Friction
\end{itemize}
\tcbline
\textbf{Skill Content:}\par
Use this skill when a ScienceWorld task asks you to compare two inclined planes by surface friction, unknown material friction, or shallowest/steepest angle, then focus on the selected inclined plane.\par

\textbf{Before the first environment action, parse and remember:}\par
\begin{itemize}
\item The two candidate plane names or descriptors exactly as the task gives them.
\item The requested target: least friction, most friction, shallowest angle, or steepest angle.
\item A single test object to use for both trials, preferably a visible brick, steel block, or orange.
\end{itemize}

\textbf{Action policy:}\par
\begin{enumerate}
\item Get to the workshop quickly. Try direct workshop navigation when available; if a movement command is rejected, use the latest visible doors or exits. Do not explore unrelated rooms once the workshop or its route is known.
\item Resolve exact plane referents before testing. Inspect or look around only enough to learn visible names such as \texttt{inclined plane d}, \texttt{inclined plane with a rubber surface}, or \texttt{inclined plane with a unknown material b surface}.
\item Pick up one available test object and use that same object on both planes. Use the stopwatch only as a helper; do not let stopwatch use create a long waiting loop.
\item For each plane, run a short comparable trial:
\begin{itemize}
\item Move or put the test object on the plane using the exact visible plane name.
\item Activate the stopwatch if it is visible and inactive.
\item Wait once with \texttt{wait}, or use at most a few \texttt{wait1} actions if fine timing is needed.
\item After the wait bound, stop waiting. Deactivate the stopwatch if it is active, pick up the same object if needed, and move to the other plane.
\end{itemize}
\item Never issue long sequences of \texttt{wait} or \texttt{wait1} during an inclined-plane comparison. After the object has clearly moved, reached the bottom, or after the bounded wait has elapsed, the trial is complete even if the observation is sparse.
\item After both planes have had one bounded trial with the same object, compare immediately:
\begin{itemize}
\item Slower movement or longer elapsed time on otherwise comparable planes means more friction.
\item Faster movement or shorter elapsed time means less friction.
\item For angle tasks with comparable surfaces and the same object, slower movement indicates the shallower angle; faster movement indicates the steeper angle.
\end{itemize}
\item If the evidence is imperfect but both planes have been sampled, choose the best-supported plane from relative motion or timing and focus. Do not gather a third trial unless a previous command was rejected and one plane was never actually tested.
\item The final action must be exactly \texttt{focus on <chosen inclined plane>} using the visible plane referent. Do not answer in natural language, do not report a score, and do not wait again once the selected plane is known.
\end{enumerate}

\textbf{Loop guards:}\par
\begin{itemize}
\item No more than one normal \texttt{wait} per plane trial.
\item No more than a few total \texttt{wait1} actions per plane trial.
\item After testing the second plane, the next non-recovery action should be the final focus command.
\item Do not use non-grammar shortcuts such as teleporting objects. If an action is rejected, inspect the visible objects briefly and retry with a shorter exact referent.
\end{itemize}
\end{skillbox}
\caption{Complete text of the Friction Skill.}
\label{skill-example-friction}
\end{figure*}

\begin{figure*}[p]
\begin{skillbox}[skillfloat]{Example: Life spans Skill}
\textbf{Skill Definition: Life spans: species lifespan comparison before life-stage focus}
\begin{itemize}
\item \textbf{Error category}: Chooses animal by visible life stage instead of comparing species life spans
\item \textbf{Applicability}: Life spans
\end{itemize}
\tcbline
\textbf{Skill Content:}\par
Use this skill for ScienceWorld tasks that ask you to find the animal with the longest life span, the shortest life span, or first the longest life span then the shortest life span. Do not use it for growing plants, life stages, object classification, genetics, friction, conductivity, temperature, or colored-box tasks.\par

\textbf{Before focusing on any animal, keep a small ledger:}\par
\begin{itemize}
\item required order: longest only, shortest only, or longest then shortest;
\item current location and route to the outside;
\item every visible outside animal referent exactly as observed;
\item normalized species for each referent after removing life-stage words such as baby, adult, juvenile, egg, young, grown, or old;
\item which required focus checkpoint has already succeeded.
\end{itemize}

\textbf{Candidate collection:}\par
\begin{enumerate}
\item First reach the outside. If a direct \texttt{go outside} is rejected, use \texttt{look around}, visible exits, and short room commands such as \texttt{go hallway}, \texttt{go greenhouse}, or \texttt{go kitchen} until an outside exit is available.
\item Once outside, use \texttt{look around} before any deliberate \texttt{focus on ...} unless the latest observation already lists the outside contents. Treat all listed animal names as candidates, including eggs and babies. Ignore tools, substances, rooms, doors, ground, fire pits, fountains, wood, and containers.
\item Do not reject eggs as non-animals in this task. A \texttt{parrot egg} is the parrot candidate, a \texttt{giant tortoise egg} is the giant tortoise candidate, and a \texttt{frog egg} is the frog or toad candidate. Focus with the exact visible referent, including \texttt{egg} or \texttt{baby}, after the species is selected.
\item If only one animal candidate is visible after looking around outside, focus that exact referent for a one-focus task. For a two-focus task, inspect outside once more or look at the candidate before deciding the second focus, but do not wander unrelated rooms.
\end{enumerate}

\textbf{Lifespan comparison policy:}\par
\begin{enumerate}
\item Compare the underlying species, not the visible life stage. Baby, egg, juvenile, grown, and adult words do not make the animal longer- or shorter-lived.
\item Use ordinary animal lifespan knowledge and broad ordering when exact years are not known:
\begin{itemize}
\item Very long-lived reptiles, especially tortoise or giant tortoise, usually outrank birds, mammals, amphibians, fish, and insects.
\item Large parrots and many turtles are generally long-lived and often outrank small mammals, amphibians, and insects.
\item Rabbits, hedgehogs, chipmunks, mice, and similar small mammals are middle or short compared with tortoises and parrots, but usually longer-lived than many small insects.
\item Frogs, toads, and chameleons are usually shorter-lived than tortoises and parrots; compare them against small mammals by best common knowledge if both appear.
\item Very short-lived insects such as ant, dragonfly, butterfly, bee, fly, and similar small insects are usually shortest among mixed candidates.
\end{itemize}
\item For a longest-only task, select the visible referent whose normalized species has the longest typical life span and execute \texttt{focus on <exact visible referent>}.
\item For a shortest-only task, select the visible referent whose normalized species has the shortest typical life span and execute \texttt{focus on <exact visible referent>}.
\item For a longest-then-shortest task, first focus the longest-lived visible species. After the observation confirms that focus, focus the shortest-lived visible species next. Do not reverse the order even when the shortest animal is more visually salient or easier to name.
\end{enumerate}

\textbf{Execution and recovery:}\par
\begin{itemize}
\item Use exact observed referents in focus commands, with the shortest unambiguous wording: \texttt{focus on giant tortoise egg}, \texttt{focus on parrot egg}, \texttt{focus on baby ant}, or \texttt{focus on baby rabbit}.
\item If a focus command produces an ambiguity menu, answer with the bare number for the intended selected species on the next action.
\item If a chosen referent is rejected or resolves to a different animal than intended, use \texttt{look around} outside or \texttt{look at <visible referent>} once, then retry with the shortest exact observed name for the same selected species.
\item Do not focus a convenient baby, egg, or first-listed animal until the species comparison has been made.
\item After all required focus checkpoints have succeeded, do not keep exploring, refocus another animal, call \texttt{task}, or answer in natural language. Let the Host or environment terminate.
\end{itemize}
\end{skillbox}
\caption{Complete text of the Life spans Skill.}
\label{skill-example-lifespans}
\end{figure*}

\begin{figure*}[p]
\begin{skillbox}[skillfloat]{Example: Object classification Skill}
\textbf{Skill Definition: Object classification: proof-first category search and target locking}
\begin{itemize}
\item \textbf{Error category}: Focused on an object that does not satisfy the requested living/plant/animal/non-living category
\item \textbf{Applicability}: Object classification
\end{itemize}
\tcbline
\textbf{Skill Content:}\par
Use this policy only for ScienceWorld tasks whose description says to find a plant, animal, living thing, or non-living thing, first focus on the thing, then move it to a colored box in a named room. For chemistry, boiling/freezing, grow-fruit, thermometer, conductivity, lifespan, inclined-plane, or other non-find-object tasks, do not apply this skill.\par

\textbf{Before the first environment action, parse and remember:}\par
\begin{itemize}
\item Requested category: plant, animal, living thing, or non-living thing.
\item Destination room and exact box color.
\item The single target after it passes the category gate. Use the same target for focus, pickup, and movement.
\end{itemize}

\textbf{Mandatory category proof before focus:}\par
\begin{itemize}
\item Do not execute \texttt{focus on <object>} until the current observation or a remembered observation proves that the same object satisfies the requested category.
\item For plant requests, valid targets are named plant organisms or visible plant parts still presented as plants: plant, tree, seedling, flowering plant, or plant/tree contents in a flower pot. Fruit, food, seeds, seed jars, soil, water, flower pots by themselves, greenhouses, tools, boxes, rooms, and animals are invalid.
\item For animal requests, valid targets are named animal organisms such as bee, bird, fish, dog, cat, parrot, mouse, tortoise, or another visible animal. Eggs, nests, containers, tools, boxes, plants, fruit, and food are invalid.
\item For living-thing requests, valid targets are actual organisms: a visible plant/tree/seedling/flowering plant or a visible animal. Fruit or food names such as orange, banana, apple, potato, peach, avocado, or berries are invalid unless the observation names them as part of a living plant target such as an apple tree. Air, agent, picture, drawing, painting, room, appliance, furniture, tool, box, jar, seed, egg, soil, water, and container are invalid.
\item For non-living-thing requests, valid targets are movable non-organism objects or substances. Do not choose the agent, a living plant, or a living animal.
\end{itemize}

\textbf{Search policy for organism categories:}\par
\begin{enumerate}
\item If a valid organism is not already visible or remembered, search before focusing. Prefer likely organism sources over destination rooms: greenhouse, flower pots, bee hive, outside, and animal containers. Use visible exits and simple commands such as \texttt{look around}, \texttt{go hallway}, \texttt{go greenhouse}, \texttt{look at flower pot <n>}, \texttt{open bee hive}, or \texttt{look in bee hive}.
\item Treat tempting invalid nouns as search cues, not targets. If the visible noun is fruit, food, air, picture, drawing, painting, box, room, container, appliance, tool, seed, egg, or agent during a plant/living/animal request, the next useful action is navigation/inspection toward organism sources, not focus or move.
\item If you already visited an ordinary room or destination room without seeing a valid organism, return to the hallway and go to greenhouse or outside instead of focusing a fallback object in that room.
\item For plant and living-thing requests, a named tree or plant in the greenhouse is a strong target. Prefer the first clear plant/tree/seedling/flowering plant over any fruit or food item.
\end{enumerate}

\textbf{Search policy for non-living categories:}\par
\begin{itemize}
\item Prefer a clearly movable visible non-organism in or near the destination room. Focus that object first, then move the same object to the requested box.
\item Avoid \texttt{wait} in find-object tasks unless an immediately preceding observation says waiting is required. After one \texttt{task} or \texttt{inventory} check, resume concrete navigation, focus, pickup, or movement.
\end{itemize}

\textbf{Execution order:}\par
\begin{enumerate}
\item Find a target that passes the category gate.
\item Execute exactly \texttt{focus on <target>} using the shortest exact visible referent. If an ambiguity menu appears, answer with the bare number for the intended valid target as the next action.
\item After focus succeeds, lock onto that target. Do not switch to a different object because it is closer or easier to move.
\item Pick up the same target when needed, or move it directly if the simulator allows direct movement.
\item Navigate to the destination room only as needed to place the locked target in the requested colored box.
\item Use \texttt{move <target> to <color> box} or \texttt{put <target> in <color> box}. If a movement ambiguity menu appears, answer the bare number for the same locked target and requested box.
\item After the observation confirms that the locked target moved to the specified box, stop manipulating objects. If another action is required before the Host ends the episode, use \texttt{task} once or \texttt{look around}; do not wait repeatedly, take the target back, refocus an invalid object, or switch targets.
\end{enumerate}

\textbf{Recovery rules:}\par
\begin{itemize}
\item If you accidentally focused an invalid object for a plant, animal, or living-thing request, do not move it. Use \texttt{look around} if the room has not been inspected, then navigate to greenhouse/outside/animal-source locations and choose a proven organism.
\item If \texttt{focus on <target>} is rejected for a plausible valid organism, inspect the current room or container once and retry with a shorter visible referent.
\item After \texttt{Unknown action} or \texttt{No known action matches}, do not repeat the same phrase or make it longer. Use \texttt{look around}, \texttt{inventory}, or a visible exit, then try a shorter command.
\item Never repeat successful focus on the same target more than once unless a later pickup or move action clearly failed and the target must be reacquired.
\end{itemize}
\end{skillbox}
\caption{Complete text of the Object classification Skill.}
\label{skill-example-classification}
\end{figure*}

\end{document}